\documentclass[letterpaper, 10 pt, conference]{ieeeconf}  % Comment this line out if you need a4paper

\IEEEoverridecommandlockouts                              % This command is only needed if 
\usepackage{graphicx} % for pdf, bitmapped graphics files
\usepackage{sdsmmTermDefs}
\usepackage{amsmath}

\usepackage{amssymb}  % assumes amsmath package installed
\usepackage{cite}
\usepackage{textcomp}

\usepackage[dvipsnames]{xcolor}
\usepackage{soul}
\setstcolor{red}

\graphicspath{ {./images/} }
\DeclareGraphicsExtensions{.pdf,.png}

\title{\LARGE \bf
Where Am I Now? Dynamically Finding Optimal Sensor States to Minimize Localization Uncertainty for a Perception-Denied Rover
}
\author{Troi Williams, Po-Lun Chen$^{\dagger}$, Sparsh Bhogavilli$^{\dagger}$, Vaibhav Sanjay$^{\dagger}$, and Pratap Tokekar% <-this % stops a space
\thanks{*This work was supported by the Computing Innovation Fellows Project 2021 (NSF Award \#2127309). All authors are at the University of Maryland, College Park, MD 20742, USA. $^\dagger$These authors did similar amounts of work.
        {\tt\small \{troiw,tokekar,pchen115\}@umd.edu}
        {\tt\small \{sbhogavi,vsanjay\}@terpmail.umd.edu}}%
}

\begin{document}

\maketitle
\thispagestyle{empty}
\pagestyle{empty}

%%%%%%%%%%%%%%%%%%%%%%%%%%%%%%%%%%%%%%%%%%%%%%%%%%%%%%%%%%%%%%%%%%%%%%%%%%%%%%%%

\begin{abstract}
	We present DyFOS, an active perception method that dynamically finds optimal states to minimize localization uncertainty while avoiding obstacles and occlusions. We consider the scenario where a perception-denied rover relies on position and uncertainty measurements from a viewer robot to localize itself along an obstacle-filled path. The position uncertainty from the viewer's sensor is a function of the states of the sensor itself, the rover, and the surrounding environment. To find an optimal sensor state that minimizes the rover's localization uncertainty, DyFOS uses a localization uncertainty prediction pipeline in an optimization search. Given numerous samples of the states mentioned above, the pipeline predicts the rover's localization uncertainty with the help of a trained, complex state-dependent sensor measurement model (a probabilistic neural network). Our pipeline also predicts occlusion and obstacle collision to remove undesirable viewer states and reduce unnecessary computations. We evaluate the proposed method numerically and in simulation. Our results show that DyFOS is faster than brute force yet performs on par. DyFOS also yielded lower localization uncertainties than faster random and heuristic-based searches.
\end{abstract}

\section{Introduction}

Navigating through known or unknown environments is challenging when an autonomous robot is ill-equipped to perceive its surrounding environment. For example, many works explore how adverse weather affects sensors typically used on autonomous cars, such as cameras, LiDAR, radar, and GPS~\cite{zang2019impact,Vargas2021Autonomous,Heinzler2019Weather,Sheeny2021RADIATE,Tang2020PerformanceTO}. Sensor performance degradation or malfunction may also occur in scenarios outside of adverse weather (for example, due to collisions). Furthermore, small robots may not have exteroceptive sensing capable of supporting robust localization. Currently, we are interested in cases where a robot cannot use its exteroceptive sensors for an extended period. Specifically, one robot (termed \emph{rover}) wants to navigate to a goal along an obstacle-filled path (Figure \ref{fig:motivation_and_high_level}). We aim to minimize the rover's localization uncertainty by leveraging cooperation from another robot (termed \emph{viewer}). Here, the rover receives position and uncertainty measurements from the viewer's sensor to localize. 

\begin{figure}[h]
    \centering
    \includegraphics[width=0.42\textwidth]{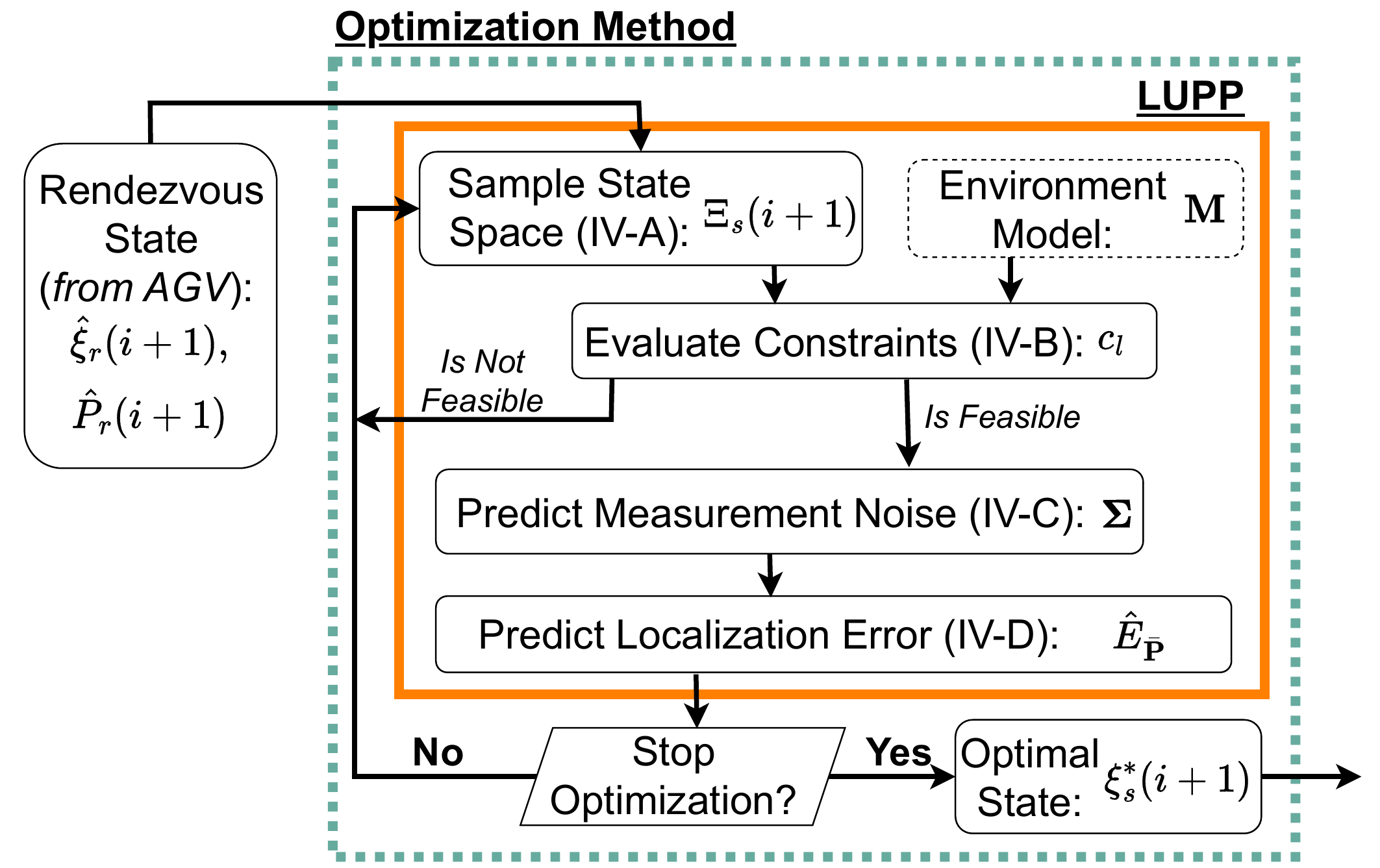}
    \caption{DyFOS is composed of an optimization method (dashed green) and LUPP (solid orange). Together, the optimizer and LUPP find an optimal sensor state $\optimalposecf{}{s}$ that has the lowest cost $\mathcal{E}_\textbf{P}$, which is derived from the rover's predicted posterior localization uncertainty at time $i+1$ (in the future). Then the viewer moves the sensor to $\optimalposecf{}{s}$ to estimate the rover's position and uncertainty. Figure \ref{fig:motivation_and_high_level} illustrates the entire state machine.}
    \label{fig:dyfos_pipeline}
\end{figure}

\begin{figure*}
\begin{center}
  \includegraphics[scale=0.3]{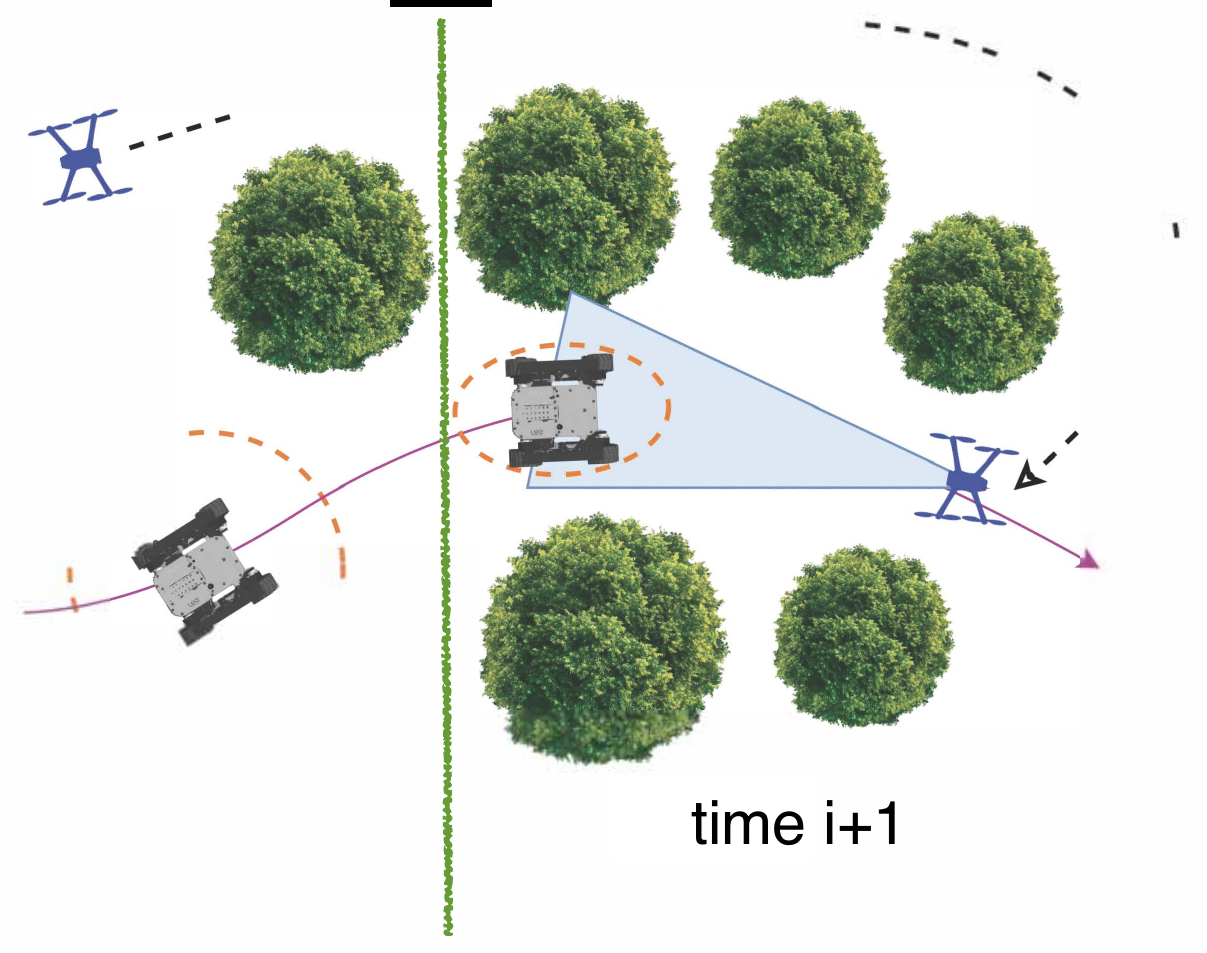}
  \includegraphics[width=0.6\textwidth]{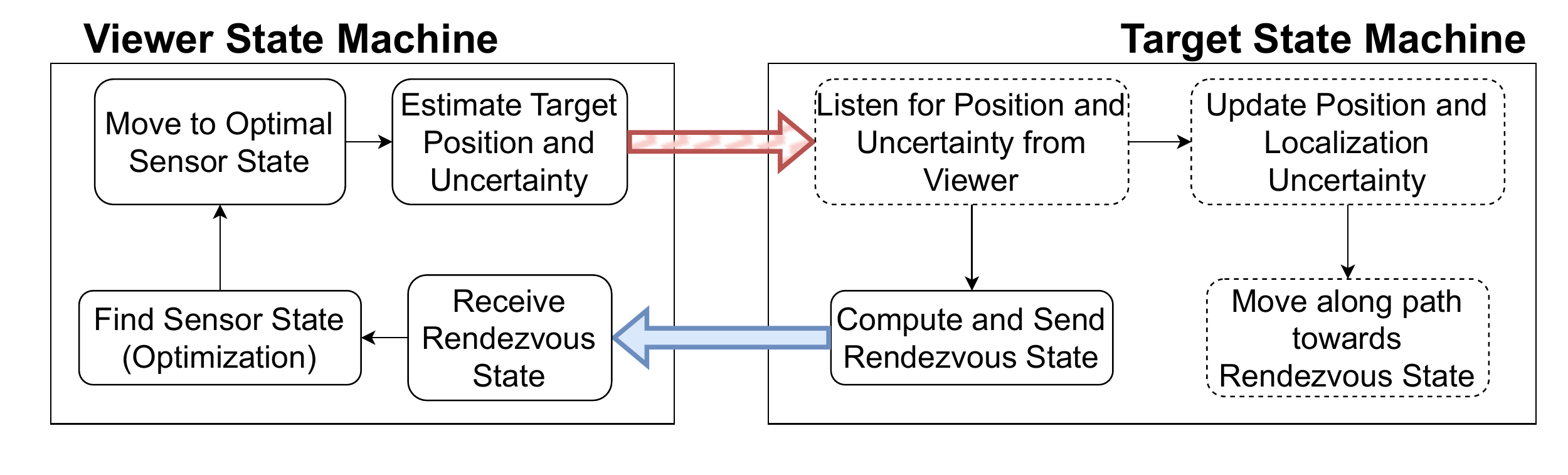}
  \caption{This figure illustrates one iteration of our motivating example (left) and a high-level representation of the viewer and rover state machines (right). The rover (grey rover) with \textit{no} exteroceptive sensors moves along a path (solid purple line) with obstacles (green trees). The rover's localization uncertainty is drawn as an orange dashed ellipse. At time $i$, the rover sends a rendezvous state to the viewer (blue drone), predicting that it (the rover) will arrive at time $i+1$. The viewer finds and moves to an optimal state that will minimize the predicted posterior localization uncertainty of the rover (see Figure \ref{fig:dyfos_pipeline} for a detailed view of the optimization module). At time $i+1$, the viewer estimates the rover's position and uncertainty and sends them to the rover. The rover updates its position and uncertainty and sends another rendezvous state to the viewer. This process repeats until the rover arrives at the goal. This paper focuses on the modules in rounded rectangles with solid lines.}
  \label{fig:motivation_and_high_level}
\end{center}
\end{figure*}

The rover's localization uncertainty is a function of the measurement uncertainty from the viewer's sensor, which is a camera in our case. We consider the scenario where the measurement uncertainty is a function of the states of the camera, rover, and surrounding environment. Although measurement uncertainty is traditionally assumed to be fixed, other works have shown that the uncertainty can be state-dependent (some include \cite{williams2019learning,williams2021learningA,Benet2002Infrared,Zhang2022tree,Huang2015Tdoa}). Therefore, we seek an optimal sensor state that avoids occlusions, avoids collisions, and has the lowest \textit{predicted}, rover localization uncertainty.

To find an optimal sensor state, we propose DyFOS, a novel active perception method that \underline{dy}namically \underline{f}inds \underline{o}ptimal \underline{s}tates to minimize the localization uncertainty of a rover. \textbf{The main contribution of DyFOS} is a localization uncertainty prediction pipeline (LUPP), a complex objective function that runs within an optimization algorithm (Figure \ref{fig:dyfos_pipeline}). Given the states of the sensor, rover, and surrounding environment, the pipeline predicts the rover's posterior localization uncertainty with the help of a state-dependent sensor measurement model (SDSMM) \cite{williams2019learning}. The pipeline also employs constraints to predict and avoid obstacle collision and occlusion. By employing the LUPP, an optimization method selects optimal sensor states that would minimize localization uncertainty and allow us to forego selecting optimal states heuristically (like in \cite{Falanga2018pampc}).

We organized the paper as follows. Section \ref{sec:related_work} discusses the related work. We formulate our problem in Section \ref{sec:problemform}. Section \ref{sec:method} describes the DyFOS algorithm. Our experiments and results are discussed in Section \ref{sec:evaluations}, and Section \ref{sec:conclude_limitations} concludes the paper.

\section{Related Work}\label{sec:related_work}

Active perception is a technique in which an agent performs a set of strategic actions that allow it to gather more insightful information about some phenomenon~\cite{bajcsy2018revisiting,aloimonos2013active,5968}. This technique has been applied to various cases including autonomous scientific information gathering \cite{Arora2019multimodal}, foreground segmentation \cite{Sun2019active}, object detection and target tracking \cite{Tokekar2011Active,Unterholzner2012active,Morbidi2013active,Price2018deep,Tallamraju2019active,Falanga2018pampc,Saini2019Markerless,Tallamraju2020AirCapRL,Zhang2022tree}, ocean flow and vehicle states \cite{Chang2022active}, and searching for individuals \cite{Acevedo2020dynamic,Sandino2020Autonomous}. For more extensive lists, Lluvia \textit{et al.} \cite{lluvia2021active} surveys methods for active mapping and robot exploration, Placed \textit{et al.} \cite{placed2022activeslam} reviews active SLAM techniques, and Queralta \textit{et al.} \cite{Queralta2020Collaborative} discusses active perception methods used in single- and multi-agent applications.

Approaches that are similar to our proposed method and motivating example include \cite{Morbidi2013active,Gurcuoglu2013Hierarchical,Falanga2018pampc,Price2018deep,Tallamraju2019active,Tallamraju2020AirCapRL}. Morbidi and Mariottini \cite{Morbidi2013active} and, subsequently, G{\"u}rc{\"u}oglu \textit{et al.} \cite{Gurcuoglu2013Hierarchical} proposed an active target tracking approach for a team of quadrotors that were equipped with 3D range sensors. In \cite{Morbidi2013active}, they explored cooperative and non-cooperative methods for minimizing the uncertainty of a moving target. They also introduced active cooperative localization and multi-target tracking, where the quadrotors fly trajectories that maximize the accuracy of their positions and the positions of multiple moving targets. In \cite{Gurcuoglu2013Hierarchical}, the authors contribute a hierarchical controller that tracks desired optimal trajectories for each quadcopter. The optimal trajectories attempt to minimize the fused position error of the target. Similar to \cite{Morbidi2013active,Gurcuoglu2013Hierarchical}, we compute optimal viewer states that minimize the perceived error of a target. However, we focus on developing a pipeline with an integrated measurement error model for scenarios within or beyond the line-of-sight of a camera.

In \cite{Falanga2018pampc}, Falanga \textit{et al.} proposed the first perception-aware, model predictive control framework for quadrotors. Their method used numerical optimization to compute trajectories that simultaneously optimized action and camera-specific, human-specified perception objectives, both of which can conflict. The human-specified perception objectives required the point of interest to be near the center of the image. Like Falanga \textit{et al.}, we optimize for perceptual objectives (namely, minimizing the localization uncertainty of a target). However, our localization uncertainty prediction pipeline (which can contain trainable components) determines the (output) viewer pose rather than heuristics. Furthermore, our method handles collision and occlusion constraints.

A line of work \cite{Price2018deep,Tallamraju2019active,Saini2019Markerless,Tallamraju2020AirCapRL} developed a cooperative detection and tracking algorithm for a team of quadrotors. The goal of these works was to track human or animal motion outdoors unconstrained and without markers. First, Price \textit{et al.} \cite{Price2018deep} showed how to achieve onboard, online, and continuous human detection and tracking in images using deep neural networks. Next, Tallamraju \textit{et al.} \cite{Tallamraju2019active} proposed an active approach to a cooperative detection and tracking algorithm (in \cite{Price2018deep}) that tracked a person performing activities. Their method minimized the 3D position error of the person by ensuring optimal viewpoint configurations of the drones. Afterward, Saini \textit{et al.} \cite{Saini2019Markerless} developed an offline method for estimating the human pose and shape using the RGB images from the aerial drone and the drone's pose, both of which were captured using \cite{Tallamraju2019active}. Finally, Tallamraju \textit{et al.} \cite{Tallamraju2020AirCapRL} proposed a deep reinforcement learning method to control the actions (formation and viewpoint configuration) of each aerial drone given a camera image. The goal of this work was to learn formation control and perceptual requirements during training and remove the need to develop hand-crafted observation models. These works either focus on detection uncertainty (instead of pose measurement uncertainty) \cite{Price2018deep,Tallamraju2019active,Saini2019Markerless} or do not model pose uncertainty explicitly \cite{Tallamraju2020AirCapRL}. They also enforce that the target is in the center of the image, which may be unnecessary for some tasks. Furthermore, although ensuring detection is vital, considering pose uncertainty too can improve localization.

\section{Problem Formulation}\label{sec:problemform}

\begin{figure}
    \centering
    \includegraphics[width=0.38\textwidth]{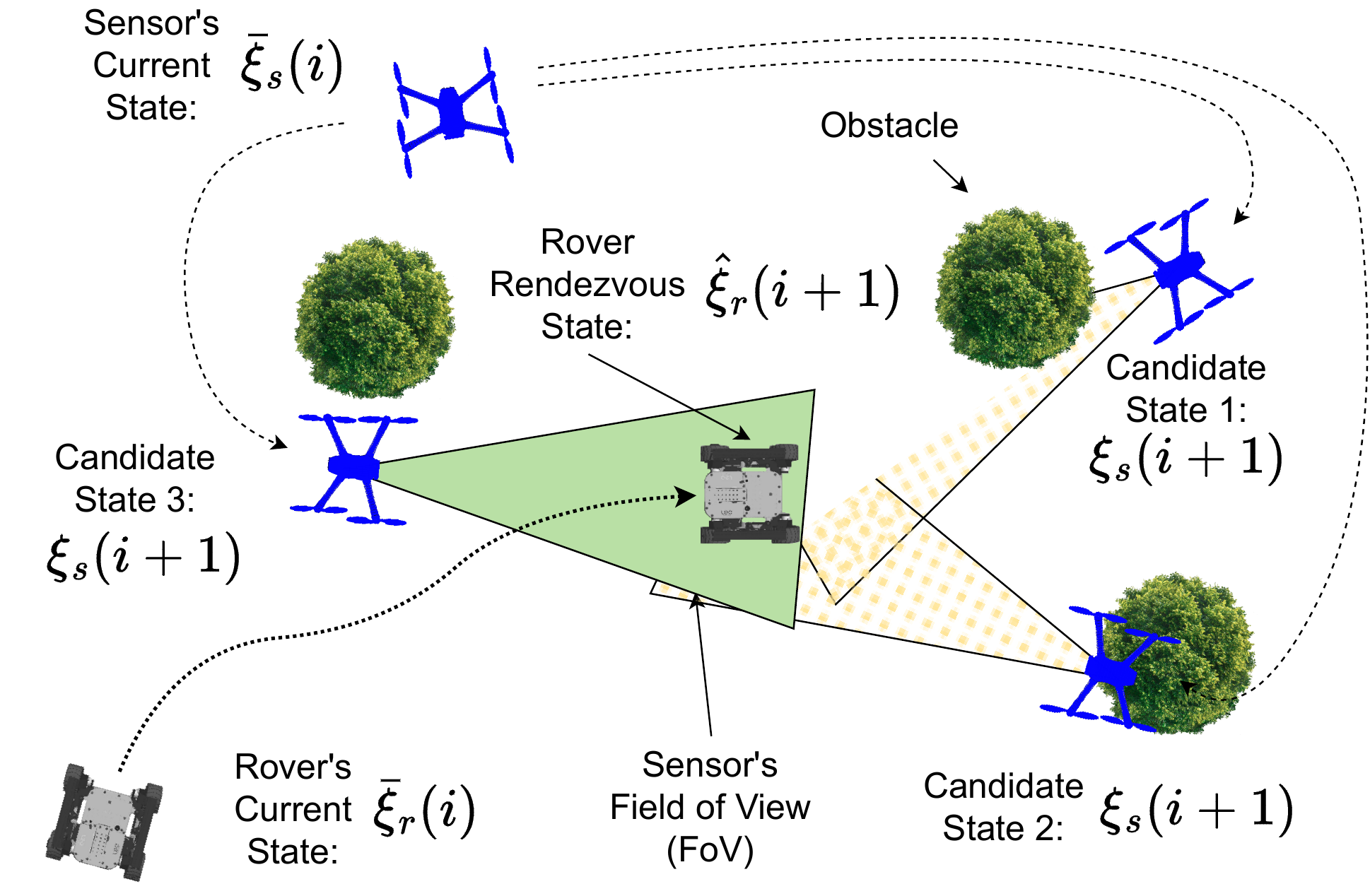}
    \caption{We show an example of predicting feasible and non-feasible sensor and viewer states via our constraints (Section \ref{sec:method:constraints}). For example, a candidate viewer state $\posecf{}{s}\vartime{i+1}$ is not feasible if we \textit{predict} the viewer will collide with obstacles (state 2) or obstacles will occlude the rover at the rendezvous state $\ppgposecf{}{r}\vartime{i+1}$ (state 1--partial FoV). Otherwise, we predict collision and occlusion will not occur (state 3 with the solid green FoV).}
    \label{fig:valid_and_invalid_viewer_states}
\end{figure}

\subsection{Preliminaries}\label{sec:problemform:prelim}

Consider a world with a planar surface, a rover, a sensor mounted on a viewer (another robot), and a collection of static obstacles (Figure \ref{fig:valid_and_invalid_viewer_states}). The rover is an autonomous ground vehicle (AGV) that wants to visit a sequence of waypoints along a path. It also has an environment model $\worldmap$ that contains information about the environment (for example, an occupancy grid and the location and direction of light sources). At time $i$, we use ${\updposecf{}{r}\vartime{i}\in\text{SE}(2)}$ to denote a posterior estimate of the rover's state and the covariance ${\updposeerrorcf{}{r}\vartime{i}\in\realNumSet{3\times3}}$ to denote the posterior state uncertainty.

Due to damage or failure of the rover's exteroceptive sensors, the rover cannot localize or avoid obstacles; it can only compute basic odometry (for example, via wheel encoders). Therefore, it relies on a viewer to help estimate its state belief. The viewer is an autonomous (aerial or ground) vehicle that can localize itself accurately. It has a copy of $\worldmap$, but does not have access to the rover's waypoints. We use ${\updposecf{}{v}\vartime{i}\in\seThree}$ to denote the estimated 6D pose of the viewer. The viewer uses a downward, front-facing camera to measure the rover's belief. The camera's estimated pose is denoted as ${\updposecf{}{s}\vartime{i}\in\seThree}$. Without loss of generality, we assume the camera is fixed to the viewer, and ${\posecf{v}{s}\in\seThree}$ describes the sensor frame relative to the viewer frame.

We define the sensor measurement (observation) model as 
\begin{equation}
    \sensorMeas\vartime{i} = \sensorModel{\mathcal{I}\vartime{i}, \updposecf{}{s}\vartime{i}, \sensorMeasNoise\vartime{i}}.
\end{equation}
Here, $\sensorModel{\cdot}$ is a vision-based algorithm that detects the rover in a camera image $\mathcal{I}\vartime{i}$ and computes the rover's state $\sensorMeas\vartime{i}\in\seTwo$ in the map frame (using the sensor's pose $\updposecf{}{s}\vartime{i}$). Due to sensor and environmental-related phenomena, $\sensorMeas\vartime{i}$ contains measurement noise $\sensorMeasNoise\vartime{i}\in\realNumSet{3}$ \cite{williams2019learning,sun2020sdsmm}. We assume \textit{each} state-dependent noise $\sensorMeasNoise\vartime{i}\sim\normal{0}{\sensorMeasNoiseCovar\vartime{i}}$ is a zero-mean Gaussian random variable with covariance $\sensorMeasNoiseCovar\vartime{i}$. Furthermore, we assume \textit{each} measurement noise covariance is state-dependent (that is, a function of the states of the viewer's sensor, rover, and environment):
\begin{equation}
    \sensorMeasNoiseCovar\vartime{i} = g(\updposecf{}{s}\vartime{i},\updposecf{}{r}\vartime{i},\worldmap),
\end{equation}
where $g(\cdot)$ is a function that outputs the state-dependent measurement noise covariance at time $i$.

At time $i$, the rover moves to the next waypoint, where it plans to rendezvous with the viewer to localize. We refer to such a waypoint as a \textit{rendezvous} (or predicted) state $\ppgposecf{}{r}\vartime{i+1}$.
Between times $i$ and $i+1$, the viewer computes an optimal sensor state and moves to such a state to observe the rover.

\subsection{Problem Statement}\label{sec:problemform:statement}

Our goal is two-fold. First, we find an optimal sensor state $\optimalposecf{}{s}\vartime{i+1}$ (in the \textit{future}) that minimizes the rover's predicted posterior localization uncertainty $\updposeerrorcf{}{r}\vartime{i+1}$ as the rover navigates along a path. Second, we seek sensor and viewer states that avoid collisions and occlusions due to obstacles. To accomplish these goals, we propose an algorithm called DyFOS, which dynamically finds an optimal data acquisition state that will 1) avoid obstacles, 2) avoid occlusions, and 3) minimize the rover's predicted posterior localization uncertainty (after a measurement update) $\updposeerrorcf{}{r}\vartime{i+1}$. We search for an optimal state $\optimalposecf{}{s}\vartime{i+1}$ using the following objective:
\begin{equation}\label{eq:optimalStateOpt}
    \begin{split}
        \optimalposecf{}{s}\vartime{i+1} & = \argmin_{\posecf{}{s}\vartime{i+1}~\in~\posesetcf{}{s}\vartime{i+1}}\optfunction{\updposeerrorcf{}{r}\vartime{i+1}} \\
        \text{s.t.} & ~\constraintfnc{l}{\updposecf{}{s}\vartime{i+1},\ppgposecf{}{r}\vartime{i+1},\worldmap} = 0, \forall l \in [1, L].
    \end{split}
\end{equation}
Here, $\ppgposecf{}{r}\vartime{i+1}$ is the rover's rendezvous state. $\posecf{}{s}\vartime{i+1}$ is a candidate sensor state that is sampled from $\posesetcf{}{s}$, the set of all sensor states. ${\constraintSet = \{\constraintsymbol{l}\}_{l=1}^L}$ are a set of constraints that predict obstacle collision or occlusion. Finally, $\worldmap$ is our environment model. Inspired by \cite{Agha2011Firm,Agha2014firm}, we choose ${\ppgposecf{}{r}\vartime{i+1}}$ as a waypoint along the rover's path. We compute ${\updposeerrorcf{}{r}\vartime{i+1}}$ by propagating the rover's current localization uncertainty $\ppgposeerrorcf{}{r}\vartime{i}$ from time $i$ to $i+1$ using the Kalman Filter equations. Finally, we assume these states can be predicted or estimated (offline and online), and, given these states, we can use an SDSMM to predict or estimate $\sensorMeasNoiseCovar$ dynamically.

\section{Dynamically Finding Optimal States to Minimize Localization Uncertainty}\label{sec:method}

The DyFOS algorithm consists of 1) an optimization method that encapsulates 2) a localization uncertainty prediction pipeline (LUPP) (Figure \ref{fig:dyfos_pipeline}). The pipeline contains four main steps. First, we sample a candidate sensor state $\posecf{}{s}$ from the sensor state space $\posesetcf{}{s}$ (Section \ref{sec:method:set_of_sensor_states}). Next, the pipeline uses a set of constraints (Section \ref{sec:method:constraints}) to determine if the current (rover, environment, and candidate sensor) state tuple is a feasible solution (for example, testing for obstacle collisions and occlusions). If the tuple is not feasible, the pipeline short-circuits and samples another candidate sensor state. However, if the state tuple is feasible, we use it to construct a feature vector $\sdsmmState$ and feed the vector to an SDSMM, which predicts the expected measurement noise covariance $\sensorMeasNoiseCovarHat$ (Section \ref{sec:method:sdsmm}). Finally, the pipeline predicts the posterior localization uncertainty $\updposeerrorcf{}{r}$ and then uses $\updposeerrorcf{}{r}$ to compute an evaluation metric (Section \ref{sec:method:localization_error}).

The optimizer executes the pipeline many times to find \textit{one} optimal sensor state $\optimalposecf{}{s}$ that produces the lowest metric value. Discussions on selecting an optimizer and finding its appropriate parameters are beyond the scope of this paper. However, Section \ref{sec:evaluations} mentions the optimizer we employed.

\subsection{The Sensor State Space}\label{sec:method:set_of_sensor_states}

The pipeline begins with sampling a candidate sensor state $\posecf{}{s}$ from the state space $\posesetcf{}{s}$. We could define $\posesetcf{}{s}$ as the set of all 3D (camera) sensor poses within a map. However, this definition leads to unnecessary evaluations because the rover (at the rendezvous pose $\ppgposecf{}{r}$) may be out of view of, too far from, or too close to the sensor for many candidate poses. Therefore, to reduce the number of (unnecessary) evaluations, we design a state space where we \textit{predict} that the rover will be within the sensor's detection range and field of view. Assuming the sensor's velocities and accelerations are near zero, we define our state space as:
\begin{equation}\label{eq:target_viewable_sensor_poses}
\begin{split}
    \posesetcf{}{s} = &~\{~\posecf{}{s}\in\text{SE}(3)~\vert~ \rho_\text{min} \leq \rho \leq \rho_\text{max},\\ &\vert\psi\vert\leq 0.5\cdot f_\text{horz},~\vert\phi\vert\leq 0.5\cdot f_\text{vert}~\}.
\end{split}
\end{equation}
The scalars $\rho_\text{min}$ and $\rho_\text{max}$ denote the camera's detection range, while the scalars $f_\text{horz}$ and $f_\text{vert}$ represent the camera's horizontal and vertical fields of view. Let $\posecf{s}{r} = [\ppgpositioncf{s}{r}, \ppgrpycf{s}{r}]^\top$ denote the rover's 3D pose (position $\ppgpositioncf{s}{r}$ and orientation $\ppgrpycf{s}{r}$) relative to the sensor. The scalar $\rho = \vert\vert\ppgpositioncf{s}{r}\vert\vert_2^2$ is the distance between the rover and the sensor. Finally, the scalars ${\psi = \arctan(\withcoordframe{s}{\hat{y}}{r}, \withcoordframe{s}{\hat{x}}{r})}$ and ${\phi = \arctan(\withcoordframe{s}{\hat{z}}{r}, \withcoordframe{s}{\hat{x}}{r})}$ (where ${\ppgpositioncf{s}{r}=[\withcoordframe{s}{\hat{x}}{r}, \withcoordframe{s}{\hat{y}}{r}, \withcoordframe{s}{\hat{z}}{r}]^\top}$) are the horizontal and vertical angles of the rover relative to the sensor, respectively.

Given the limits of $\rho$, $\phi$, and $\psi$ in \eqref{eq:target_viewable_sensor_poses}, we can independently sample values for each variable to create our state space $\posesetcf{}{s}$, the set of 3D camera poses where we predict the rover is detectable and within the camera's field of view.

\subsection{Sensor and Viewer State Constraints}\label{sec:method:constraints}

During the online search, we use constraints to \textit{predict} if a candidate state is feasible beyond the line-of-sight of the sensor (Figure \ref{fig:valid_and_invalid_viewer_states}). Let ${\constraintSet = \{\constraintsymbol{_l}\}_{l=1}^L}$ contain a set of constraints that determines if a sampled state ${\posecf{}{s}\in\posesetcf{}{s}}$ is a feasible solution. Collectively, the set of constraints $\constraintSet$ defines a feasible region that varies with time and space, and this feasible region contains the sensor state we seek for optimal data acquisition. In addition, the constraints prevent the pipeline from running the SDSMM and computing the posterior localization uncertainty if $\posecf{}{s}$ is not feasible, reducing unnecessary computation.

We use two constraints and a 3D occupancy grid (such as an Octomap \cite{hornung13octomap}) to determine if $\posecf{}{s}$ is feasible. The first constraint predicts collisions between the viewer (robot) and obstacles by emitting fixed-length rays horizontally and vertically from $\posecf{}{v}$. Here, we assume $\posecf{v}{s}$, the transform from the sensor to the vehicle frame, is known. The second constraint predicts if obstacles will occlude the rover. To predict occlusions, we emit rays from $\posecf{}{s}$ to points along the perimeter of the rover at $\ppgposecf{}{r}$, where the perimeter is a generic bounding shape. For either constraint, $\posecf{}{s}$ is \textit{not} feasible if \textit{any} ray intersects with a voxel within the occupancy grid.

\subsection{State-Dependent Sensor Measurement Model}\label{sec:method:sdsmm}

\begin{figure}
    \centering
    \includegraphics[clip, trim=8cm 0cm 8cm 0cm, width=0.44\textwidth]{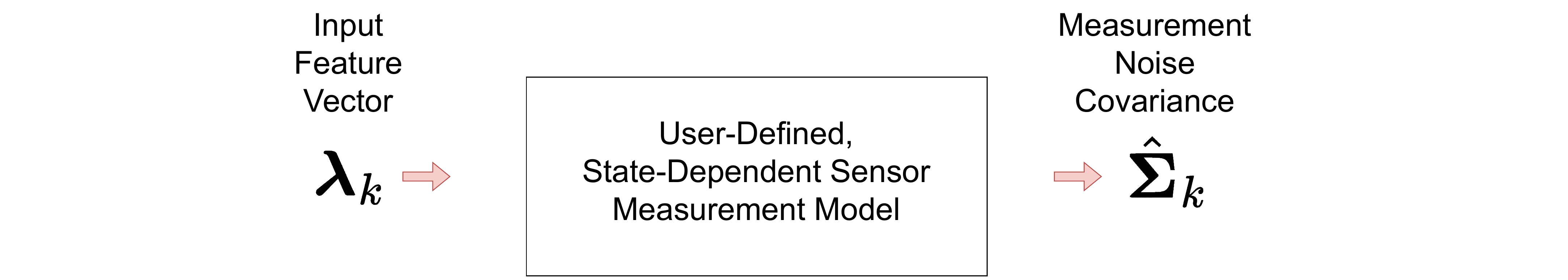}
    \caption{We use the states of the viewer, rover, and surrounding environment to construct a feature vector $\sdsmmState$. Then, given $\sdsmmState$, the SDSMM outputs a covariance of the measurement noise $\sensorMeasNoiseCovarHat$. We describe how to implement an SDSMM using neural networks (such as in \cite{sun2020sdsmm,williams2019learning,williams2021learningA}) in Section \ref{sec:method:sdsmm}.}
    \label{fig:sample_mdn}
\end{figure}

\subsubsection{Overview} An SDSMM is a model that outputs a state-dependent measurement distribution \cite{williams2019learning,sun2020sdsmm} (Figure \ref{fig:sample_mdn}). We can use such distributions to derive statistics such as the measurement bias and noise. Examples of an SDSMM include linear models, multi-layer perceptions, or Gaussian Processes. The input to an SDSMM is a set of features $\sdsmmState$ that correlate with the measurement bias and noise\footnote{Williams and Sun \cite{williams2019learning,sun2020sdsmm} refer to such features as a combined state. However, we avoid using the term to alleviate confusion with robot states.}, and, in our case, the output is the predicted or estimated covariance of the measurement noise $\sensorMeasNoiseCovarHat$ of the rover's position.  For our experiments, we implement an SDSMM as a fixed linear model. However, we describe how to learn an SDSMM in the context of neural networks below (such as in \cite{williams2019learning,sun2020sdsmm}).

\subsubsection{Training an SDSMM} We train an SDSMM using a dataset ${\dataset = \{(\sensorMeasError\vartime{k}, \sdsmmState\vartime{k})\}}_{k=1}^K$ that contains $K$ measurement errors and their corresponding features. The measurement error is defined as ${\sensorMeasError\vartime{k} = (\sensorMeas\vartime{k} - \posecf{}{r}\vartime{k})}$, where $\sensorMeas\vartime{k}$ is a measurement of the rover's pose using a pose estimation algorithm and $\posecf{}{r}\vartime{k}$ is the ground truth pose of the rover. Generally speaking, the feature vector $\sdsmmState\vartime{k}$ for a given sensor and estimation problem varies. For instance, a feature vector for a camera used in a rover localization problem may contain the ambient lighting of the area and the distance and relative velocity between the camera and the rover. We note that these features must be calculable using the ground truth, estimated, or predicted of the states of the sensor, rover, and surrounding environment. For example, we use the ground truth and estimated states to compute the features during training. However, during runtime, we use the estimated or predicted states to compute the features. Section \ref{sec:evaluations} describe the SDSMM and features used in our experiments.

Given the dataset $\dataset$, we use the negative log-likelihood loss $\lossfncsymbol$ to train our model. Since our measurement noise is normally-distributed, we simplify the loss to
\begin{equation}\label{eq:negloglikeloss_simple}
   \lossfncsymbol = -\frac{1}{2}\sum_{k = 1}^K\left(\ln{(\vert\sensorMeasNoiseCovarHat\vartime{k}\vert)} + \sensorMeasError\vartime{k}^\top[\sensorMeasNoiseCovarHat\vartime{k}]^{-1}\sensorMeasError\vartime{k}\right),
\end{equation}
where $\sensorMeasNoiseCovarHat\vartime{k}$ is the output covariance from the SDSMM.

\subsubsection{Dual Usage} In this paper, the SDSMM has two purposes. First, while evaluating candidate sensor poses (via the LUPP), the SDSMM \textit{predicts} the measurement noise covariance using predicted features derived from a candidate sensor pose, rover's rendezvous pose, and environment (OctoMap). This capability allows us to predict the measurement noise beyond the sensor's line of sight. In the second case (when the sensor is at the optimal pose), the SDSMM \textit{estimates} the measurement noise covariance using features estimated from the current sensor, rover, and environment states. Then we transmit the estimated noise covariance to the rover (as shown in Figure \ref{fig:motivation_and_high_level}).

\subsection{Posterior Localization Uncertainty Prediction}\label{sec:method:localization_error}

The final step in our pipeline predicts a posterior localization uncertainty (\textit{after} a future measurement update) $\updposeerrorcf{}{r}$ and its associated metric value $\mathcal{E}_\textbf{P}$. Following Gürcüoğlu \textit{et al.} \cite{Gurcuoglu2013Hierarchical}, we use the following function
\begin{equation}\label{eq:pred_localization_error}
    \mathcal{E}_\textbf{P} = \optfunction{\updposeerrorcf{}{r}} =  \ln\left(\det\left(\updposeerrorcf{}{r}\right)\right).
\end{equation}
Since we use a Kalman Filter to track the rover's position and uncertainty, we predict $\updposeerrorcf{}{r}$ using the Kalman Filter Riccati equations (like in \cite{Gurcuoglu2013Hierarchical,Zhang2022tree}):
\begin{equation}
    \updposeerrorcf{}{r}=\left\{\textbf{I} - \left(\ppgposeerrorcf{}{r}\textbf{H}^\top\textbf{S}_t^{-1}\ppgposeerrorcf{}{r}\right)\textbf{H}\right\}\ppgposeerrorcf{}{r},
\end{equation}
where $\textbf{I}\in\realNumSet{3\times 3}$ is an identity matrix, ${\textbf{S}_t = \textbf{H}\ppgposeerrorcf{}{r}\textbf{H}^\top + \withcoordframe{}{\sensorMeasNoiseCovarHat}{r}}$ is the innovation covariance, $\ppgposeerrorcf{}{r}$ is the rover's \textit{a priori} localization uncertainty (propagated from time $i$ to $i+1$), and $\withcoordframe{}{\sensorMeasNoiseCovarHat}{r}$ is the predicted measurement uncertainty (using the SDSMM and a set of \textit{predicted} features). Finally, we highlight that $\updposeerrorcf{}{r}$ and $\mathcal{E}_\textbf{P}$ also vary with each candidate sensor state because $\sensorMeasNoiseCovarHat$ is a function of the sensor state.

Once the pipeline computes $\mathcal{E}_\textbf{P}$ for a candidate state $\posecf{}{s}$, the optimizer checks if it should continue searching for an optimal state or terminate. If the pipeline terminates, it either performed the maximum number of iterations or did not improve over some period. In either case, we set the optimal sensor state $\optimalposecf{}{s}$ as the state with the lowest cost. Then the viewer moves the sensor to $\optimalposecf{}{s}$.

\section{Evaluations}\label{sec:evaluations}

We performed numerical and simulated evaluations to compare our proposed method. In each evaluation, we compared DyFOS with the random and center view baseline algorithms, where each algorithm computed a pose for the viewer's sensor. The random algorithm arbitrarily placed the viewer and is akin to using a traditional, state-\textit{independent} measurement noise model with a constant variance. Center view placed the viewer such that the rover is in the center of the image. This algorithm implicitly assumes measurement noise varies and is lowest when the rover is in the center of an image (this heuristic is similar to \cite{Falanga2018pampc}). Finally, DyFOS used the LUPP (Section \ref{sec:method}) with a differential evaluation global optimizer \cite{storn1997differential} in SciPy \cite{2020SciPy-NMeth}. In our numerical evaluation, we also used a brute force algorithm with the LUPP to find the ``best possible'' sensor pose. All algorithms employed the constraints in Section \ref{sec:method:constraints}. We also used belief propagation to compute the rendezvous belief.

In DyFOS and brute force, the LUPP used an SDSMM to predict and estimate the measurement noise covariance. We implemented our SDSMM using a linear model for both evaluations. The 2D position noise model was defined as ${\sensorMeasNoiseCovar = \text{diag}(\sigma_x, \sigma_y)}$, where $\sigma_x$ and $\sigma_y$ have the same value: ${\sigma_{\ast} = \sigma_1^2 + \sigma_2^2 + (c\cdot\sigma_3)^2}$. Here, $\sigma_1^2 = 0.03^2$ was the minimum measurement noise. The term $\sigma_2^2 \in [0.03^2, 0.3^2]$ represented the noise due to the location of the rover in the camera image, where $0.03^2$ was used if the rover was in the center and $0.3^2$ was used if the rover was closer to the edge. Finally, $(c\cdot\sigma_3)^2$ was the measurement noise due to the size of the rover in the image (which correlates with distance) and a simulated reflection scalar $c$ due to the sun, where $\sigma_3\in[0.03, 0.5]$. We used the law of reflection to compute $c\in[0, 2]$, where $0$ means no specular reflection and $2$ means the sun's reflection partially obscured the rover.

When executing the pipeline, the SDSMM predicted the measurement noise covariance given predictions of the reflected sunlight $c$ and the marker's predicted size and location within a camera image. We predicted these features (the intensity and the marker's size and location in an image) using the predicted states (that is, the rover's rendezvous belief, the viewer's candidate pose, and the position of the sun). We estimated the measurement noise covariance using the estimated corresponding states whenever the viewer observed the rover (in the simulated evaluation).

\subsection{Numerical Evaluation}\label{sec:evaluation:numerical}

\subsubsection{Overview} We generated $20$ arbitrary 2D maps, where each map had one rover and $25$ static obstacles. We randomized the poses for the rover and obstacles. The diameter of the viewer and rover were $0.75$ meters and $1.5$ meters, respectively. We sampled the diameter of each obstacle using the uniform distribution $\mathcal{U}(0.5, 2.5)$. For each map, we computed a random, \textit{a priori} localization uncertainty ${\ppgposeerrorcf{}{r}\vartime{i+1} = \mathbf{A}^\top\mathbf{A}}$, where $\mathbf{A}\in\realNumSet{2\times 2}$ is randomized. Finally, we implemented our constraints using Shapely \cite{Sean2007Shapely}.

\begin{table}
    \centering
    \begin{tabular}{|c||c|c|c|}
        \hline
              & Random View & Center View & DyFOS (proposed) \\
         \hline\hline
         RMSE & $26.8$      & $41.3$      & $\boldsymbol{7.6}$ \\
         \hline
       Median & $17.9$      & $36.6$      & $\boldsymbol{5.1}$ \\
         \hline
         Mean & $22.7$      & $36.2$      & $\boldsymbol{6.5}$ \\
         Stdv.& $14.3$      & $20.0$      & $\boldsymbol{4.0}$ \\
         \hline
         Max  & $53.8$      & $82.3$      & $\boldsymbol{16.4}$\\
        \hline
    \end{tabular}
    \caption{Statistics describing the absolute translation error (in cm) of the rover in the simulated evaluations. Best is \textbf{bolded}.}
    \label{table:simulation_ape}
\end{table}

\subsubsection{Results} For each algorithm, we evaluated this experiment using the criteria below.

\begin{figure}
    \centering
    \includegraphics[clip, scale=0.28]{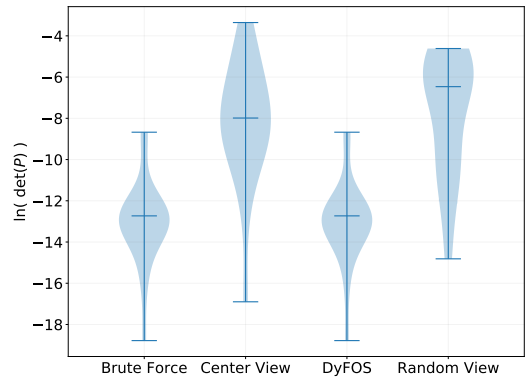}
    \caption{The numerical experiments show that DyFOS (proposed) performs on par with the brute force approach and outperforms the quicker random and center (heuristic) view searches. The y-axis is the D-optimal measure of the localization uncertainty $\ppgposeerrorcf{}{r}\vartime{i+1}$. For each violin (algorithm), the top and bottom notches are the extrema, and the middle notch is the median.}
    \label{fig:numerical_lndetloss}
\end{figure} 

\textbf{Predicted Posterior Localization Uncertainty.} Aside from brute force, DyFOS had the lowest values, performing on par with the brute force algorithm while evaluating fewer candidate poses (see Figure \ref{fig:numerical_lndetloss}). Moreover, the worst case for DyFOS was still projected to perform better than at least $50\%$ of the random and center view results. Interestingly, the performance of the center and random view algorithms varied more than DyFOS and brute force. We believe the higher variance in performance is due to the less predictable nature of both algorithms. We discuss this notation more in our simulated results.

\textbf{Run Times}: The run times varied due to the complexity of each algorithm. The random and center view algorithms were the quickest (less than $0.01$ s each) since they did not use the pipeline to find a viewer pose. The run time for DyFOS was $1.87\pm0.58$ s with a maximum of $3.09$ s using one core, which may be reasonable for real-time usage if the algorithm is optimized. Finally, the brute force algorithm took $41.62\pm6.02$ s with a maximum of $46.95$ s using all $8$ cores to find a view pose.

\subsection{Simulated Experiment}\label{sec:evaluation:simulated}

\begin{figure}
    \centering
    \includegraphics[clip, trim=6cm 0.5cm 0cm 1.28cm, scale=0.55]{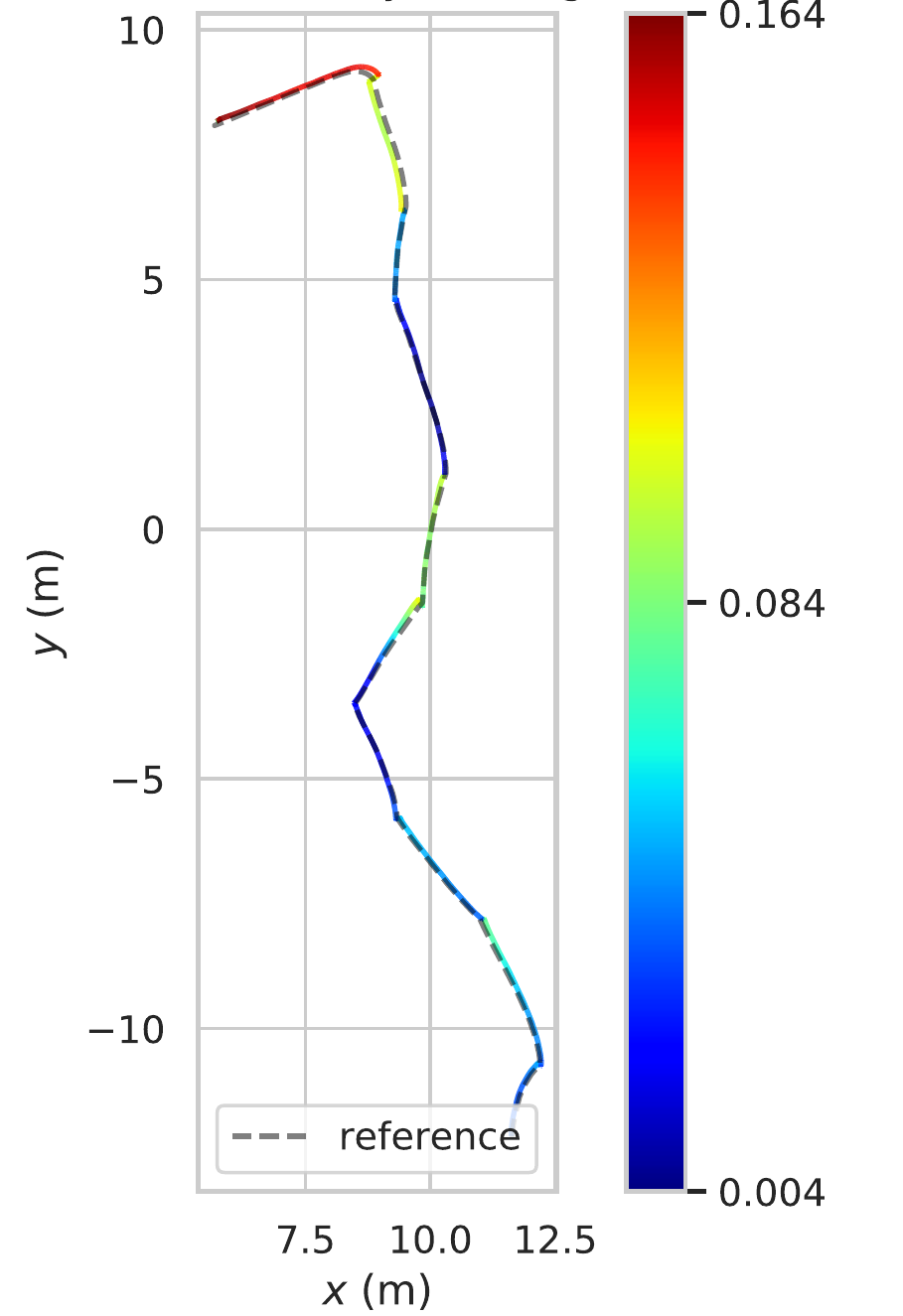}
    \caption{This figure shows the actual, or reference, path (in dashed lines) and the estimated path (in solid gradient colors) of the rover when the viewer used DyFOS to select viewpoints. The color bar to the right shows the range of localization errors for the rover during the experiment.}
    \label{fig:dyfos_ape}
\end{figure}

\begin{figure}
    \centering
    \includegraphics[scale=0.34]{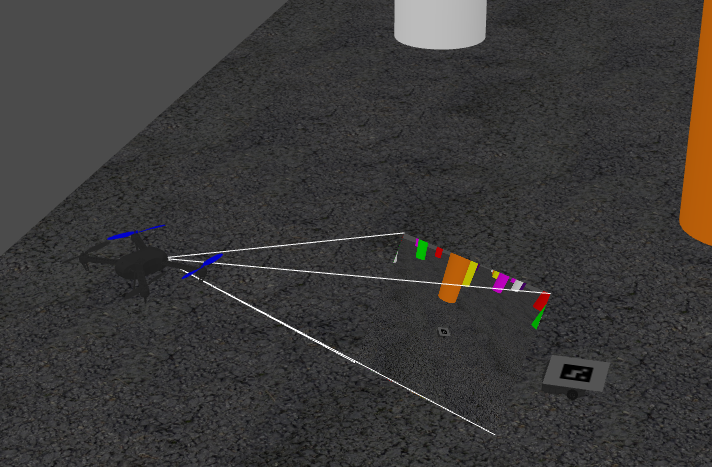}
    \caption{The simulated experiments in Gazebo were performed in a world with many obstacles (colored cylinders). The drone (viewer) localized the Turtlebot3 (rover) with the help of the ArUco marker mounted on its top.}
    \label{fig:sim_exp_setup}
\end{figure}

\subsubsection{Overview} The simulated experiments were performed in the Gazebo simulator \cite{koenig2004gazebo}; Figure \ref{fig:sim_exp_setup} depicts part of the simulated world. The Gazebo world was a planar surface, and it contained $25$ static obstacles that were $10$ meters tall, had a diameter of one meter, and were in the workspaces of the viewer and rover. Both robots used the state machines (like in Figure \ref{fig:motivation_and_high_level}) to navigate and communicate. An OctoMap \cite{hornung13octomap} was used to implement our constraints.

We performed one experiment for each algorithm. In each experiment, the rover visited $10$ fixed waypoints with a total distance of over $20$ m (see Figure \ref{fig:dyfos_ape}).

\subsubsection{Results} We discuss the simulated results below.

\textbf{Rover Pose Error.} Our results showed that DyFOS also had the lowest absolute pose error (APE) (see Table \ref{table:simulation_ape}). We believe DyFOS achieved the lowest APE because it actively searches for viewpoints that will reduce the rover's localization uncertainty with the help of the SDSMM. This results in behaviors such as looking away from the sun to remove reflections. Random view was the second-best performer, mainly due to its random behavior. For example, sometimes random view may choose a good viewpoint with low noise, and other times it may choose poorly. Finally, the center view algorithm performed the worst. The poor performance may be due to the camera's angle and not considering other sources of measurement noise (for example, sunlight reflections). At the camera's current angle, the viewer must be at least $0.8$ m away from the rover so that it appears in the middle of the image, which results in higher amounts of noise. Furthermore, since center view does not consider other sources of noise, this algorithm (like random view) may select viewpoints with high amounts of noise, even though it is satisfying its heuristic.

\section{Conclusion}\label{sec:conclude_limitations}

We proposed DyFOS, a novel active perception method that \underline{Dy}namically \underline{F}inds \underline{O}ptimal \underline{S}tates to minimize localization uncertainty while avoiding obstacles and occlusions. DyFOS relies on user-defined constraints, a learned state-dependent sensor measurement model, and an objective function to evaluate the quality of each candidate state with respect to the localization task. DyFOS exploits an SDSMM to minimize the rover's predicted posterior localization uncertainty in a hazard-filled world. Our experiments showed that DyFOS performs on par with brute force methods but is significantly faster. The average runtime was under two seconds using one CPU core, which may be reasonable for some real-time applications if we optimized the algorithm. DyFOS also achieves lower predicted localization uncertainties than random and heuristic search algorithms.

\addtolength{\textheight}{-1cm}   % This command serves to balance the column lengths
                                  % on the last page of the document manually. It shortens
                                  % the textheight of the last page by a suitable amount.
                                  % This command does not take effect until the next page
                                  % so it should come on the page before the last. Make
                                  % sure that you do not shorten the textheight too much.

\bibliographystyle{unsrt}

\begin{thebibliography}{10}

    \bibitem{zang2019impact}
    Shizhe Zang, Ming Ding, David Smith, Paul Tyler, Thierry Rakotoarivelo, and
      Mohamed~Ali Kaafar.
    \newblock {The Impact of Adverse Weather Conditions on Autonomous Vehicles: How
      Rain, Snow, Fog, and Hail Affect the Performance of a Self-Driving Car}.
    \newblock {\em IEEE Vehicular Technology Magazine}, 14(2):103--111, 2019.
    
    \bibitem{Vargas2021Autonomous}
    Jorge Vargas, Suleiman Alsweiss, Onur Toker, Rahul Razdan, and Joshua Santos.
    \newblock {An Overview of Autonomous Vehicles Sensors and Their Vulnerability
      to Weather Conditions}.
    \newblock {\em Sensors}, 21(16), 2021.
    
    \bibitem{Heinzler2019Weather}
    Robin Heinzler, Philipp Schindler, Jürgen Seekircher, Werner Ritter, and
      Wilhelm Stork.
    \newblock {Weather Influence and Classification with Automotive Lidar Sensors}.
    \newblock In {\em 2019 IEEE Intelligent Vehicles Symposium (IV)}, pages
      1527--1534, 2019.
    
    \bibitem{Sheeny2021RADIATE}
    Marcel Sheeny, Emanuele De~Pellegrin, Saptarshi Mukherjee, Alireza Ahrabian,
      Sen Wang, and Andrew Wallace.
    \newblock {RADIATE: A Radar Dataset for Automotive Perception in Bad Weather}.
    \newblock In {\em 2021 IEEE International Conference on Robotics and Automation
      (ICRA)}, pages 1--7, 2021.
    
    \bibitem{Tang2020PerformanceTO}
    Li~Tang, Yunpeng Shi, Qing He, Adel~W. Sadek, and Chunming Qiao.
    \newblock {Performance Test of Autonomous Vehicle Lidar Sensors Under Different
      Weather Conditions}.
    \newblock {\em Transportation Research Record}, 2674:319 -- 329, 2020.
    
    \bibitem{williams2019learning}
    Troi Williams and Yu~Sun.
    \newblock {Learning State-Dependent, Sensor Measurement Models for
      Localization}.
    \newblock In {\em 2019 IEEE/RSJ International Conference on Intelligent Robots
      and Systems (IROS)}, pages 3090--3097, 2019.
    
    \bibitem{williams2021learningA}
    Troi Williams and Yu~Sun.
    \newblock {Learning State-Dependent Sensor Measurement Models with Limited
      Sensor Measurements}.
    \newblock In {\em 2021 IEEE/RSJ International Conference on Intelligent Robots
      and Systems (IROS)}, pages 86--93, 2021.
    
    \bibitem{Benet2002Infrared}
    G.~Benet, F.~Blanes, J.E. Simó, and P.~Pérez.
    \newblock {Using infrared sensors for distance measurement in mobile robots}.
    \newblock {\em Robotics and Autonomous Systems}, 40(4):255--266, 2002.
    
    \bibitem{Zhang2022tree}
    Zhongshun Zhang and Pratap Tokekar.
    \newblock {Tree Search Techniques for Adversarial Target Tracking With
      Distance-Dependent Measurement Noise}.
    \newblock {\em IEEE Transactions on Control Systems Technology},
      30(2):712--727, 2022.
    
    \bibitem{Huang2015Tdoa}
    Baoqi Huang, Lihua Xie, and Zai Yang.
    \newblock {TDOA-Based Source Localization With Distance-Dependent Noises}.
    \newblock {\em IEEE Transactions on Wireless Communications}, 14(1):468--480,
      2015.
    
    \bibitem{Falanga2018pampc}
    Davide Falanga, Philipp Foehn, Peng Lu, and Davide Scaramuzza.
    \newblock {PAMPC: Perception-Aware Model Predictive Control for Quadrotors}.
    \newblock In {\em 2018 IEEE/RSJ International Conference on Intelligent Robots
      and Systems (IROS)}, pages 1--8, 2018.
    
    \bibitem{bajcsy2018revisiting}
    Ruzena Bajcsy, Yiannis Aloimonos, and John~K Tsotsos.
    \newblock {Revisiting Active Perception}.
    \newblock {\em Autonomous Robots}, 42(2):177--196, 2018.
    
    \bibitem{aloimonos2013active}
    Yiannis Aloimonos.
    \newblock {\em {Active Perception}}.
    \newblock Psychology Press, 2013.
    
    \bibitem{5968}
    R.~Bajcsy.
    \newblock {Active Perception}.
    \newblock {\em Proceedings of the IEEE}, 76(8):966--1005, 1988.
    
    \bibitem{Arora2019multimodal}
    Akash Arora, P.~Michael Furlong, Robert Fitch, Salah Sukkarieh, and Terrence
      Fong.
    \newblock {Multi-Modal Active Perception for Information Gathering in Science
      Missions}.
    \newblock {\em Auton. Robots}, 43(7):1827–1853, oct 2019.
    
    \bibitem{Sun2019active}
    Yuxiang Sun, Ming Liu, and Max Q.-H. Meng.
    \newblock {Active Perception for Foreground Segmentation: An RGB-D Data-Based
      Background Modeling Method}.
    \newblock {\em IEEE Transactions on Automation Science and Engineering},
      16(4):1596--1609, 2019.
    
    \bibitem{Tokekar2011Active}
    Pratap Tokekar, Joshua Vander~Hook, and Volkan Isler.
    \newblock {Active Target Localization for Bearing Based Robotic Telemetry}.
    \newblock In {\em 2011 IEEE/RSJ International Conference on Intelligent Robots
      and Systems}, pages 488--493, 2011.
    
    \bibitem{Unterholzner2012active}
    Alois Unterholzner, Michael Himmelsbach, and Hans-Joachim Wuensche.
    \newblock {Active Perception for Autonomous Vehicles}.
    \newblock In {\em 2012 IEEE International Conference on Robotics and
      Automation}, pages 1620--1627, 2012.
    
    \bibitem{Morbidi2013active}
    Fabio Morbidi and Gian~Luca Mariottini.
    \newblock {Active Target Tracking and Cooperative Localization for Teams of
      Aerial Vehicles}.
    \newblock {\em IEEE Transactions on Control Systems Technology},
      21(5):1694--1707, 2013.
    
    \bibitem{Price2018deep}
    Eric Price, Guilherme Lawless, Roman Ludwig, Igor Martinovic, Heinrich~H.
      Bülthoff, Michael~J. Black, and Aamir Ahmad.
    \newblock {Deep Neural Network-Based Cooperative Visual Tracking Through
      Multiple Micro Aerial Vehicles}.
    \newblock {\em IEEE Robotics and Automation Letters}, 3(4):3193--3200, 2018.
    
    \bibitem{Tallamraju2019active}
    Rahul Tallamraju, Eric Price, Roman Ludwig, Kamalakar Karlapalem, Heinrich~H.
      Bülthoff, Michael~J. Black, and Aamir Ahmad.
    \newblock {Active Perception Based Formation Control for Multiple Aerial
      Vehicles}.
    \newblock {\em IEEE Robotics and Automation Letters}, 4(4):4491--4498, 2019.
    
    \bibitem{Saini2019Markerless}
    Nitin Saini, Eric Price, Rahul Tallamraju, Raffi Enficiaud, Roman Ludwig, Igor
      Martinovic, Aamir Ahmad, and Michael Black.
    \newblock {Markerless Outdoor Human Motion Capture Using Multiple Autonomous
      Micro Aerial Vehicles}.
    \newblock In {\em 2019 IEEE/CVF International Conference on Computer Vision
      (ICCV)}, pages 823--832, 2019.
    
    \bibitem{Tallamraju2020AirCapRL}
    Rahul Tallamraju, Nitin Saini, Elia Bonetto, Michael Pabst, Yu~Tang Liu,
      Michael~J. Black, and Aamir Ahmad.
    \newblock {AirCapRL: Autonomous Aerial Human Motion Capture Using Deep
      Reinforcement Learning}.
    \newblock {\em IEEE Robotics and Automation Letters}, 5(4):6678--6685, 2020.
    
    \bibitem{Chang2022active}
    Dongsik Chang, Matthew Johnson-Roberson, and Jing Sun.
    \newblock {An Active Perception Framework for Autonomous Underwater Vehicle
      Navigation Under Sensor Constraints}.
    \newblock {\em IEEE Transactions on Control Systems Technology}, pages 1--16,
      2022.
    
    \bibitem{Acevedo2020dynamic}
    José~J. Acevedo, João Messias, Jesús Capitán, Rodrigo Ventura, Luis Merino,
      and Pedro~U. Lima.
    \newblock {A Dynamic Weighted Area Assignment Based on a Particle Filter for
      Active Cooperative Perception}.
    \newblock {\em IEEE Robotics and Automation Letters}, 5(2):736--743, 2020.
    
    \bibitem{Sandino2020Autonomous}
    Juan Sandino, Fernando Vanegas, Felipe Gonzalez, and Frederic Maire.
    \newblock {Autonomous UAV Navigation for Active Perception of Targets in
      Uncertain and Cluttered Environments}.
    \newblock In {\em 2020 IEEE Aerospace Conference}, pages 1--12, 2020.
    
    \bibitem{lluvia2021active}
    Iker Lluvia, Elena Lazkano, and Ander Ansuategi.
    \newblock {Active Mapping and Robot Exploration: A Survey}.
    \newblock {\em Sensors}, 21(7), 2021.
    
    \bibitem{placed2022activeslam}
    Julio~A. Placed, Jared Strader, Henry Carrillo, Nikolay Atanasov, Vadim
      Indelman, Luca Carlone, and José~A. Castellanos.
    \newblock {A Survey on Active Simultaneous Localization and Mapping: State of
      the Art and New Frontiers}.
    \newblock 2022.
    
    \bibitem{Queralta2020Collaborative}
    Jorge~Peña Queralta, Jussi Taipalmaa, Bilge Can~Pullinen, Victor~Kathan
      Sarker, Tuan Nguyen~Gia, Hannu Tenhunen, Moncef Gabbouj, Jenni Raitoharju,
      and Tomi Westerlund.
    \newblock {Collaborative Multi-Robot Search and Rescue: Planning, Coordination,
      Perception, and Active Vision}.
    \newblock {\em IEEE Access}, 8:191617--191643, 2020.
    
    \bibitem{Gurcuoglu2013Hierarchical}
    Utku Gürcüoglu, Gustavo~A. Puerto-Souza, Fabio Morbidi, and Gian~Luca
      Mariottini.
    \newblock {Hierarchical Control of a Team of Quadrotors for Cooperative Active
      Target Tracking}.
    \newblock In {\em 2013 IEEE/RSJ International Conference on Intelligent Robots
      and Systems}, pages 5730--5735, 2013.
    
    \bibitem{sun2020sdsmm}
    Yu~Sun and Troi Williams.
    \newblock Learning state-dependent sensor measurement models for localization,
      February 2020.
    \newblock USF Patents.
    
    \bibitem{Agha2011Firm}
    Ali-akbar Agha-mohammadi, Suman Chakravorty, and Nancy~M. Amato.
    \newblock {FIRM: Feedback controller-based information-state roadmap - A
      framework for motion planning under uncertainty}.
    \newblock In {\em 2011 IEEE/RSJ International Conference on Intelligent Robots
      and Systems}, pages 4284--4291, 2011.
    
    \bibitem{Agha2014firm}
    Ali-Akbar Agha-Mohammadi, Suman Chakravorty, and Nancy~M Amato.
    \newblock {FIRM: Sampling-based feedback motion-planning under motion
      uncertainty and imperfect measurements}.
    \newblock {\em The International Journal of Robotics Research}, 33(2):268--304,
      2014.
    
    \bibitem{hornung13octomap}
    Armin Hornung, Kai~M. Wurm, Maren Bennewitz, Cyrill Stachniss, and Wolfram
      Burgard.
    \newblock {OctoMap: An Efficient Probabilistic 3D Mapping Framework Based on
      Octrees}.
    \newblock {\em Autonomous Robots}, 2013.
    \newblock Software available at https://octomap.github.io.
    
    \bibitem{storn1997differential}
    Rainer Storn and Kenneth Price.
    \newblock {Differential Evolution--A Simple and Efficient Heuristic for Global
      Optimization over Continuous Spaces}.
    \newblock {\em Journal of global optimization}, 11(4):341--359, 1997.
    
    \bibitem{2020SciPy-NMeth}
    Pauli Virtanen, Ralf Gommers, Travis~E. Oliphant, Matt Haberland, Tyler Reddy,
      David Cournapeau, Evgeni Burovski, Pearu Peterson, Warren Weckesser, Jonathan
      Bright, St{\'e}fan~J. {van der Walt}, Matthew Brett, Joshua Wilson, K.~Jarrod
      Millman, Nikolay Mayorov, Andrew R.~J. Nelson, Eric Jones, Robert Kern, Eric
      Larson, C~J Carey, {\.I}lhan Polat, Yu~Feng, Eric~W. Moore, Jake
      {VanderPlas}, Denis Laxalde, Josef Perktold, Robert Cimrman, Ian Henriksen,
      E.~A. Quintero, Charles~R. Harris, Anne~M. Archibald, Ant{\^o}nio~H. Ribeiro,
      Fabian Pedregosa, Paul {van Mulbregt}, and {SciPy 1.0 Contributors}.
    \newblock {{SciPy} 1.0: Fundamental Algorithms for Scientific Computing in
      Python}.
    \newblock {\em Nature Methods}, 17:261--272, 2020.
    
    \bibitem{Sean2007Shapely}
    Sean Gillies et~al.
    \newblock {Shapely: Manipulation and Analysis of Geometric Objects}, 2007--.
    
    \bibitem{koenig2004gazebo}
    N.~Koenig and A.~Howard.
    \newblock Design and use paradigms for gazebo, an open-source multi-robot
      simulator.
    \newblock In {\em 2004 IEEE/RSJ International Conference on Intelligent Robots
      and Systems (IROS) (IEEE Cat. No.04CH37566)}, volume~3, pages 2149--2154
      vol.3, 2004.
    
    \end{thebibliography}

\end{document}